\documentclass[10pt,conference]{IEEEtran}
\IEEEoverridecommandlockouts
\usepackage{cite}
\usepackage{amsmath,amssymb,amsfonts}
\usepackage{algorithm}
\usepackage{algpseudocode}
\usepackage{booktabs}
\usepackage{array}
\usepackage{float}
\usepackage{graphicx}
\usepackage{textcomp}
\usepackage{xcolor}
\usepackage{tikz}
\usepackage{hyperref} 
\usetikzlibrary{arrows.meta, positioning, fit, calc, backgrounds}
\def\BibTeX{{\rm B\kern-.05em{\sc i\kern-.025em b}\kern-.08em
    T\kern-.1667em\lower.7ex\hbox{E}\kern-.125emX}}
\begin{document}

\title{Importance-Aware Scheduling for High-Dimensional Hyperparameter Optimization}
\author{
\IEEEauthorblockN{Ruinan Wang, Ian Nabney, Mohammad Golbabaee}
\IEEEauthorblockA{University of Bristol, Bristol, United Kingdom}
\IEEEauthorblockA{\texttt{\{zg21696, in17746, an22148\}@bristol.ac.uk}}
}
\maketitle

\begin{abstract}
Hyperparameter Optimization (HPO) is essential for building high-performing ML/DL models, yet conventional optimizers often struggle in high-dimensional spaces where evaluations are costly and progress is diluted across many low-impact variables. We propose Greedy Importance First (GIF), an importance-aware scheduling strategy that uses a small-sample warm start to estimate hyperparameter importance, forms importance-based groups, allocates trials proportionally, and retains a full-space fallback. We evaluate GIF under fixed evaluation budgets on five anisotropic analytic functions ($d\!\in\!\{5,10,30,50\}$), Bayesmark, and NAS-Bench-301 (33D). On the higher-dimensional benchmarks, GIF reaches better incumbents with faster convergence than TPE, BOHB, Random Search, and Sequential Grouping. On Bayesmark, where the effective dimensionality is smaller, GIF remains competitive but the margins are smaller. Ablation studies show that importance estimation, proportional allocation, and the fallback step all contribute to the gains. We also verify that the HIA component recovers the intended anisotropy on the analytic benchmarks. These results suggest that GIF is a simple and plug-compatible way to improve sample efficiency in high-dimensional HPO.

\end{abstract}


\section{Introduction}
Hyperparameter optimization (HPO) is a critical stage in modern ML/DL pipelines: it governs robustness, stability, and generalization. Despite a mature toolbox---Bayesian optimization (e.g., TPE~\cite{bergstra2011algorithms}, BOHB~\cite{falkner2018bohb}), evolutionary~\cite{loshchilov2016cma}, and bandit methods~\cite{li2018hyperband}---efficiency often degrades as dimensionality grows: each evaluation becomes costlier and surrogates become harder to fit and less informative~\cite{bischl2023hyperparameter}. Crucially, the obstacle is not dimensionality alone but the strongly uneven influence of hyperparameters~\cite{probst2019tunability}. In many models, a small subset of settings accounts for most performance variation, while others contribute marginally. Yet most optimizers advance all coordinates in lockstep each iteration, effectively enforcing uniform scheduling. This induces a dimensionality bottleneck: treating all hyperparameters equally dilutes the budget and delays progress, especially under tight evaluation limits.

Hyperparameter importance assessment (HIA) provides a principled foundation for addressing this bottleneck: from a small set of trials, it estimates each hyperparameter's marginal contribution to performance---and, when needed, pairwise interactions. However, despite the availability of estimators such as N-RReliefF~\cite{wang2024efficient}, fANOVA~\cite{hutter2014efficient}, and PED-ANOVA~\cite{watanabe2023ped}, there is no widely adopted strategy that operationalizes these estimates into concrete scheduling decisions. As a result, HIA methods are underutilized in practice.

\begin{figure}[t]
\centering
\resizebox{\columnwidth}{!}{%
\begin{tikzpicture}[
  >=Latex,
  node distance=1.2cm and 1.4cm,
  font=\Large,
  nodebox/.style={
    draw, rounded corners=6pt, line width=0.6pt,
    fill=white, align=center, inner sep=6pt, text width=4.6cm
  },
  flowarrow/.style={-Latex, line width=1.6pt, draw=blue!70}
]

\node[nodebox] (pilot) {\textbf{Warm start optimization}};
\node[nodebox, right=of pilot] (hia) {\textbf{HIA}};
\node[nodebox, right=of hia] (group) {\textbf{Importance-sorted grouping}};

\node[nodebox, below=1.1cm of group] (alloc) {\textbf{Allocate trials by group importance weights}};
\node[nodebox, below=1.1cm of alloc] (groupopt) {\textbf{Group-wise optimization}};

\node[nodebox, below=4.63cm of hia] (fallback)
{\textbf{Full-space fallback if no gain}};

\draw[flowarrow] (pilot) -- (hia);
\draw[flowarrow] (hia) -- (group);
\draw[flowarrow] (group) -- (alloc);
\draw[flowarrow] (alloc) -- (groupopt);

\draw[flowarrow] (groupopt.west) -- (fallback.east);
\draw[flowarrow] (fallback.north) -- (hia.south);

\node[draw=black, rounded corners=12pt, line width=0.6pt, dashed,
      fit=(pilot)(hia)(group)(alloc)(groupopt)(fallback),
      inner sep=10pt] (innerframe) {};

\node[nodebox, below=1.2cm of innerframe.south] (outputs)
{\textbf{Outputs}\\
\emph{$(\mathbf{h_{\textbf{best}},y_{\textbf{best}}})$} and \\
\emph{$(\mathbf{H},\mathbf{Y})$}};

\draw[flowarrow] (innerframe.south) -- (outputs.north);

\begin{scope}[on background layer]
  \node[draw=black, rounded corners=14pt, line width=0.6pt, fill=blue!6,
        fit=(innerframe)(outputs),
        inner sep=14pt] (outerframe) {};
\end{scope}

\end{tikzpicture}%
}
\caption{\textbf{GIF Pipeline:} High-level workflow of the proposed Greedy Importance First strategy.}
\label{fig:gif-pipeline-layout}
\end{figure}
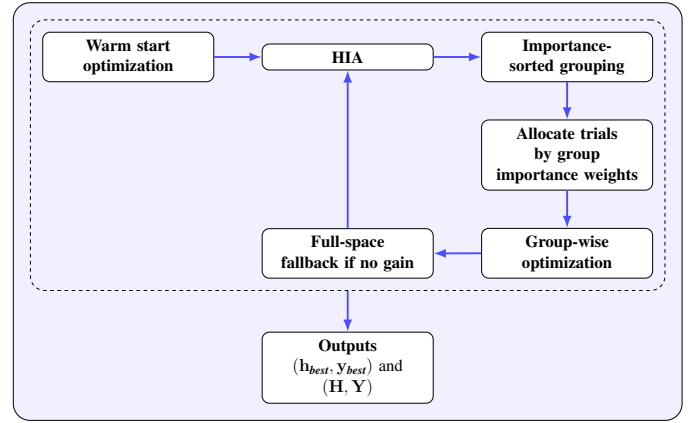

This paper introduces Greedy Importance-First (GIF), an importance-aware HPO strategy that turns HIA insights into an explicit, budgeted search plan. As illustrated in Fig.~\ref{fig:gif-pipeline-layout}, GIF (i) performs a small-sample warm start to collect an initial history for HIA HIA algorithms; (ii) orders hyperparameters by estimated importance and groups them accordingly; (iii) allocates budgets proportionally to group importance and optimizes each group while fixing other variables at the current incumbent, warm-starting from the accumulated history; and (iv) when a round yields no improvement, falls back to joint optimization to restore global exploration. This design concentrates the budget where it matters most, while the fallback to joint optimization provides a principled escape from local stagnation. We evaluate GIF under fixed budgets on controlled anisotropic analytic functions, Bayesmark tasks~\cite{bayesmark2019}, and NAS-Bench-301~\cite{zela2020surrogate}. Ablations disentangle the effect of each component, and we further verify that HIA can recover the ground-truth anisotropy on the analytic benchmarks—even with limited evaluations, it correctly highlights the few dominant coordinates while suppressing negligible ones.

\textbf{Contributions:}
\begin{itemize}
\item We propose Greedy Importance First (GIF), which converts HIA estimates into a concrete search plan. Specifically, GIF orders hyperparameters by estimated importance, partitions them into groups, allocates trials proportionally, and introduces a per-round full-space fallback. It provides a simple pathway to more economical high-dimensional HPO.
\item We validate that standard HIA estimators (e.g., N\,-RReliefF) recover ground-truth anisotropy on controlled functions, supporting their use as reliable priors when budgets are tight.
\item Under fixed evaluation budgets and multiple random seeds, GIF consistently outperforms various established HPO baselines on higher-dimensional benchmarks (five anisotropic analytic functions, NAS-Bench-301), while remaining competitive on lower/mid-dimensional Bayesmark tasks (4 models $\times$ 5 datasets). In high-dimensional scenarios, GIF achieves a markedly better accuracy--time trade-off.
\item Ablation studies show that the introduced components---(i) importance-driven ranking, (ii) proportional budget allocation, and (iii) the full-space joint fallback---each contributes to the overall gains; removing any single component leads to a clear drop in performance.
\end{itemize}

\section{Related Work}
\textbf{Hyperparameter Importance Assessment (HIA).} Understanding which hyperparameters “matter” has long supported post-hoc analysis and space design; for example, Weights \& Biases (W\&B) Sweeps provide importance plots from trial histories~\cite{wandb_sweeps}, while libraries such as Optuna and SMAC3 expose fANOVA-based importance tools~\cite{akiba2019optuna, lindauer2022smac3}. Methodologically, fANOVA remains a standard variance-decomposition approach~\cite{hutter2014efficient}; PED-ANOVA generalizes it with a Pearson-divergence–based closed form that enables efficient local importance on arbitrary subspaces (e.g., top-performing regions)~\cite{watanabe2023ped}. Complementary to fANOVA-style decompositions, N-RReliefF adapts ReliefF to continuous responses and quantifies both marginal and pairwise interaction importance from HPO histories, offering a lightweight, data-driven estimator under tight budgets~\cite{wang2024efficient}.

\textbf{Gray-box and uncertainty-aware HPO.} Gray-box approaches enrich BO surrogates with intermediate training signals (e.g., learning curves, checkpoint features, or partial-fidelity measurements), and uncertainty-aware schedulers couple candidate selection with budget allocation to avoid premature discarding under early-stage noise~\cite{liu2024uq, mehta2024improving, falkner2018bohb}. While these methods exploit richer signals across candidates, GIF serves as a lightweight allocator across hyperparameters: it uses HIA from small warm-starts to reweight search effort across hyperparameters and can plug into TPE-style optimizers as the inner engine.

\textbf{Resource allocation, warm starts, and scheduling.} Many HPO systems exploit warm starts (e.g., transferring priors or surrogate states), parallel scheduling, or multi-fidelity allocation across candidates and tasks~\cite{wistuba2018scalable, falkner2018bohb, li2018hyperband, swersky2013multi, wang2025grouped}. However, they typically retain uniform treatment across hyperparameters within an iteration. GIF breaks this per-iteration uniformity by fixing non-targeted hyperparameters to the current incumbent and concentrating trials on the most important groups. This increases the signal-to-noise ratio per evaluation in high-dimensional regimes. A per-round full-space fallback then provides a principled escape hatch from local plateaus.

In sum, GIF turns early HIA into a concrete search plan: importance-ordered grouping, importance-proportional allocation, warm-started subspace search, and a safeguarded full-space fallback. This yields a plug-compatible route to sample-efficient HPO in high-dimensional settings, complementary to structural high-dimensional BO (subspace/variable-selection assumptions and local/trust-region BO), gray-box surrogates, and uncertainty-aware schedulers.

\section{Problem Setup}
\label{sec:problem-setup}

We consider HPO on a fixed dataset \(\mathcal{D}\) and hyperparameter search space \(\Theta=\Theta_1\times\cdots\times\Theta_d\), where each \(\Theta_i\) is the domain of hyperparameter \(H_i\). A configuration is \(\mathbf{h}=(h_1,\ldots,h_d)\in\Theta\). The black-box objective is $f_{\mathcal{D}}:\Theta\to\mathbb{R}, \mathbf{h}\mapsto f_{\mathcal{D}}(\mathbf{h}),$ which returns a scalar performance (e.g., validation accuracy to maximize). The goal of HPO is $\mathbf{h}^\star \in \arg\max_{\mathbf{h}\in\Theta} f_{\mathcal{D}}(\mathbf{h}),  y^\star = f_{\mathcal{D}}(\mathbf{h}^\star),$ subject to a limited evaluation (or wall-clock) budget \(B_{\text{total}}\). After \(t\) evaluations, the history is \(\mathcal{H}=\{\mathbf{h}^{(1)},\ldots,\mathbf{h}^{(t)}\}\) and \(\mathcal{Y}=\{y^{(1)},\ldots,y^{(t)}\}\) with \(y^{(i)}=f_{\mathcal{D}}(\mathbf{h}^{(i)})\). The incumbent (best-so-far) configuration is \((h_{\text{best}},y_{\text{best}})\) where \(y_{\text{best}}=\max_{i\le t} y^{(i)}\). An optimizer \(\mathcal{A}_{\text{opt}}\) proposes new candidates conditioned on \((\mathcal{H},\mathcal{Y})\), evaluates them, and appends results until \(B_{\text{total}}\) is exhausted. Standard outputs are the final incumbent \((h_{\text{best}},y_{\text{best}})\) and the complete trace \((\mathcal{H},\mathcal{Y})\).

\textbf{Representative baseline.}
TPE~\cite{bergstra2011algorithms} partitions the history \((\mathcal{H},\mathcal{Y})\) by a score threshold \(y_0\) (e.g., the \(\gamma\)-quantile), and fits conditional densities \(l(\mathbf{h})=p(\mathbf{h}\mid y\ge y_0)\) and \(g(\mathbf{h})=p(\mathbf{h}\mid y< y_0)\). New candidates maximize \(l(\mathbf{h})/g(\mathbf{h})\), a proxy for expected improvement. In practice, \(l\) and \(g\) are estimated via Parzen windows with the factorization \(p(\mathbf{h})\approx \prod_{j=1}^d p(h_j)\). The iterative loop is: fit densities \(\to\) sample \(\mathbf{h}^{(t+1)}\) \(\to\) evaluate \(f_{\mathcal{D}}(\mathbf{h}^{(t+1)})\) \(\to\) update the history.

\textbf{Typical bottlenecks in High Dimensions.} For the representative optimizer TPE, as the dimensionality \(d\) of the search space increases, several limitations arise under tight evaluation budgets \(B_{\text{total}}\): 
\textbf{(i)} The independence assumption \(p(\mathbf{h}) \approx \prod_{j} p(h_{j})\) neglects coordinate interactions. In high-dimensional hyperparameter spaces, many variables only matter through their joint effects. Ignoring such dependencies causes both \(l(\mathbf{h})\) and \(g(\mathbf{h})\) to appear nearly uniform across most coordinates, offering little guidance for exploration.
\textbf{(ii)} In higher dimensions, density estimation becomes increasingly noisy because the effective sample size per coordinate shrinks. With limited evaluations, each marginal distribution is poorly supported, so \(l(\mathbf{h})\) and \(g(\mathbf{h})\) fluctuate heavily, yielding unstable search guidance. 
\textbf{(iii)} As \(d\) increases, contributions of coordinates to \(f\) are highly imbalanced; the presence of many low-impact dimensions reduces the effective signal-to-noise in each sampled evaluation, leading to slower improvement over iterations.
These effects help explain why standard BO methods such as TPE often struggle in high-dimensional settings. GIF addresses this issue by using hyperparameter importance estimates to order variables, form groups, and allocate evaluations more selectively.

\section{The GIF Algorithm}

\subsection{Pipeline Overview}
\label{subsec:pipeline}
Algorithm~\ref{alg:gif} formalizes how the key components of GIF are orchestrated into a single scheduling strategy.

\begin{algorithm}[t]
\caption{GIF Main Strategy}
\label{alg:gif}
\begin{algorithmic}[1]
\Require Search space $\Theta$, objective $f_{\mathcal{D}}$, subsample ratio $\alpha$,
initial budget $B_{\mathrm{init}}$, step size $B_{\mathrm{step}}$, total budget $B_{\mathrm{total}}$,
max group size $k$, importance evaluator $\mathcal{A}_{\mathrm{imp}}$, optimizer $\mathcal{A}_{\mathrm{opt}}$, fallback ratio $\rho$
\Ensure Incumbent $(\mathbf{h}_{\mathrm{best}},y_{\mathrm{best}})$, complete evaluation trace $(\mathcal{H},\mathcal{Y})$

\State $(\mathcal{H},\mathcal{Y}), (\mathbf{h}_{\mathrm{best}},y_{\mathrm{best}}) \gets$
\Statex \hspace*{\algorithmicindent}\textbf{WarmStart}$(\Theta, f_{\mathcal{D}}, \alpha, B_{\mathrm{init}}, \mathcal{A}_{\mathrm{opt}})$ 

\State $T_{\mathrm{used}} \gets B_{\mathrm{init}}$, \quad
       $T_{\mathrm{full\,used}} \gets 0$, \quad
       $B_{\mathrm{full\,total}} \gets \rho\, B_{\mathrm{total}}$

\While{$T_{\mathrm{used}} < B_{\mathrm{total}}$}

  \State $I \gets \mathcal{A}_{\mathrm{imp}}(\mathcal{H}, \mathcal{Y})$
  \Comment{importance weights $\{I_i\}_{i=1}^d$}

  \State \textbf{FormGroups:} sort indices by $I$ (desc.), then partition into groups
        $\mathcal{G}=\{\mathcal{G}_j\}$ with $|\mathcal{G}_j|\le k$

  \State $B_{\mathrm{cur}} \gets \min\!\big(B_{\mathrm{step}},\, B_{\mathrm{total}}-T_{\mathrm{used}}\big)$

  \State $\mathbf{b} \gets \textbf{AllocateBudget}(\mathcal{G}, I, B_{\mathrm{cur}})$

  \State $(\mathcal{H},\mathcal{Y}, \mathbf{h}_{\mathrm{best}}, y_{\mathrm{best}}, T_{\mathrm{used}}, improved) \gets$
  \Statex \hspace*{\algorithmicindent}\textbf{GroupOpt}$(\mathcal{G}, \mathbf{b}, \mathcal{H}, \mathcal{Y},
        \mathbf{h}_{\mathrm{best}}, y_{\mathrm{best}}, T_{\mathrm{used}}, \mathcal{A}_{\mathrm{opt}})$

  \If{\textbf{not} $improved$ \textbf{and} $T_{\mathrm{full\,used}} < B_{\mathrm{full\,total}}$
      \textbf{and} $T_{\mathrm{used}} < B_{\mathrm{total}}$}

    \State $R \gets \left\lfloor \tfrac{B_{\mathrm{total}}-T_{\mathrm{used}}}{B_{\mathrm{step}}} \right\rfloor + 1$

    \State $B_{\mathrm{full}} \gets$
     \Statex \hspace*{\algorithmicindent}$
      \min\!\left(\left\lfloor \tfrac{B_{\mathrm{full\,total}}-T_{\mathrm{full\,used}}}{R} \right\rfloor,\,
      B_{\mathrm{total}}-T_{\mathrm{used}}\right)$

\State $(\mathcal{H},\mathcal{Y}, \mathbf{h}_{\mathrm{best}}, y_{\mathrm{best}}, T_{\mathrm{used}}, T_{\mathrm{full\,used}}) \gets$
\Statex \hspace*{\algorithmicindent}\textbf{FullSpaceOpt}$(\Theta, f_{\mathcal{D}}, B_{\mathrm{full}},
    \mathcal{H}, \mathcal{Y}, \mathbf{h}_{\mathrm{best}}, y_{\mathrm{best}},
    T_{\mathrm{used}},$
\Statex \hspace*{\algorithmicindent}\hspace*{1.2em}%
    $T_{\mathrm{full\,used}}, \mathcal{A}_{\mathrm{opt}})$

  \EndIf

\EndWhile

\State \Return $(\mathbf{h}_{\mathrm{best}},y_{\mathrm{best}})$ and $(\mathcal{H},\mathcal{Y})$
\end{algorithmic}
\end{algorithm}

\subsection{Warm Start}
\label{subsec:warm-start}
\textbf{Inputs:} Search space $\Theta$, dataset $\mathcal{D}$ (size $|\mathcal{D}|$), objective $f_{\mathcal{D}}:\Theta\to\mathbb{R}$, subsample ratio $\alpha\in(0,1]$, warm-start budget $B_{\mathrm{init}}$, inner optimizer $\mathcal{A}_{\mathrm{opt}}$. \textbf{Outputs:} Initial history $(\mathcal{H},\mathcal{Y})$ with $|\mathcal{H}|=|\mathcal{Y}|=B_{\mathrm{init}}$, and incumbent $(\mathbf{h}_{\mathrm{best}},y_{\mathrm{best}})$.
We randomly subsample the dataset to obtain $\mathcal{D}_{\mathrm{init}}$ of size $\alpha|\mathcal{D}|$ and run $\mathcal{A}_{\mathrm{opt}}$ for $B_{\mathrm{init}}$ evaluations on $\Theta$ (using $\mathcal{D}_{\mathrm{init}}$), producing $(\mathcal{H},\mathcal{Y})$ and initializing $(\mathbf{h}_{\mathrm{best}},y_{\mathrm{best}})$ as the best in this history. The warm-start stage reduces early evaluation cost while providing a more informative history for subsequent importance estimation than purely random initialization.

\subsection{Hyperparameter Importance Assessment (HIA)}
\label{subsec:hia}
\textbf{Inputs:} Optimization history $(\mathcal{H},\mathcal{Y})$; evaluator $\mathcal{A}_{\mathrm{imp}}$. \textbf{Outputs:} Normalized importance profile $\{I_i\}_{i=1}^d$ assigning a nonnegative weight to each hyperparameter $H_i$.

In general, HIA methods assign weights $I_1,\ldots,I_d$ estimating each hyperparameter’s marginal contribution to performance, providing interpretable insights about “what matters” and informing downstream scheduling or search-space design. 
Representative techniques include \emph{fANOVA}~\cite{hutter2014efficient}, \emph{PED-ANOVA}~\cite{watanabe2023ped}, and \emph{N-RReliefF}~\cite{wang2024efficient}.
In GIF, we employ \emph{N-RReliefF} as our default importance evaluator. Given history $(\mathcal{H},\mathcal{Y})$, N-RReliefF treats each configuration as a reference point, compares it with its nearest neighbors in configuration space, and accumulates per-dimension covariation weighted by the performance difference between neighbors. This produces raw scores $\widehat{I}_i$, which are then mapped into positive, comparable importances via a softplus normalization and re-scaled so that $\sum_i I_i=1$. In this way, dimensions where small input changes consistently lead to large performance shifts are assigned higher weights, which in turn underpin ordering and grouping in GIF.

\subsection{Grouping and Allocation}
\label{subsec:groupalloc}
\textbf{Key Inputs:} Importance weights $\{I_i\}_{i=1}^d$; maximum group size $k$; per-round step size $B_{\mathrm{step}}$; total budget $B_{\mathrm{total}}$; used trials $T_{\mathrm{used}}$. 
\textbf{Outputs:} A partition of hyperparameter indices into groups $\mathcal{G}=\{\mathcal{G}_j\}$ with $|\mathcal{G}_j|\le k$, and per-group trials allocations $\mathbf{b}=[b_1,\ldots,b_{|\mathcal{G}|}]$.
We first set the current round budget $B_{\mathrm{cur}} \;=\; \min\!\bigl(B_{\mathrm{step}},\; B_{\mathrm{total}}-T_{\mathrm{used}}\bigr).$ Based on $\{I_i\}$, we sort hyperparameters by descending weight and partition them into groups of size at most $k$. For each group $\mathcal{G}_j$, we compute its total weight $I_j=\sum_{i\in\mathcal{G}_j} I_i$ and allocate trials proportionally: $b_j=\max\!\left(1,\ \Bigl\lfloor \tfrac{I_j}{\sum_{g}I_g}\,B_{\mathrm{cur}}\Bigr\rfloor\right).$ Enforcing $b_j\ge 1$ guarantees at least one trial per group; the final allocations $\mathbf{b}$ are then passed to the group-wise optimization stage. If rounding leaves unassigned trials, we distribute the remaining trials one by one to the groups with the largest fractional remainders, ensuring that the final integer allocations exactly match the per-round budget.

\subsection{Group-wise Optimization}
\label{subsec:groupopt}
\textbf{Key Inputs:} Groups $\mathcal{G}=\{\mathcal{G}_j\}$, per-group allocations $\mathbf{b}=[b_1,\ldots,b_{|\mathcal{G}|}]$, 
current history $(\mathcal{H},\mathcal{Y})$, incumbent $(\mathbf{h}_{\mathrm{best}},y_{\mathrm{best}})$, and inner optimizer $\mathcal{A}_{\mathrm{opt}}$. 
\textbf{Outputs:} Updated history $(\mathcal{H},\mathcal{Y})$, updated incumbent $(\mathbf{h}_{\mathrm{best}},y_{\mathrm{best}})$, 
and updated trial counter $T_{\mathrm{used}}$.

For each group $\mathcal{G}_j$, we fix all hyperparameters outside $\mathcal{G}_j$ to their values in the current incumbent $\mathbf{h}_{\mathrm{best}}$. 
We then invoke the inner optimizer $\mathcal{A}_{\mathrm{opt}}$ for $b_j$ evaluations restricted to $\mathcal{G}_j$, with \texttt{warm-start} from the existing history $(\mathcal{H},\mathcal{Y})$. 
The resulting evaluations $(\mathcal{H}_{j},\mathcal{Y}_{j})$ are appended to the history, and $T_{\mathrm{used}}$ is incremented by $b_j$. 
After each group is optimized, we update the incumbent if a better configuration is discovered. 
If all groups fail to improve the incumbent, the round is considered \emph{unsuccessful}, potentially triggering the full-space fallback. 
Otherwise, the algorithm proceeds with the next round using the updated history and incumbent.

\subsection{Full-Space Fallback}
\label{subsec:fallback}
\textbf{Key Inputs:} Remaining trials $T_{\mathrm{left}} = B_{\mathrm{total}} - T_{\mathrm{used}}$; full-space reserved quota $B_{\mathrm{full\,total}}=\rho\,B_{\mathrm{total}}$; cumulative full-space trials used $T_{\mathrm{full\,used}}$ (i.e., trials already spent on full-space fallback); step size $B_{\mathrm{step}}$. \textbf{Outputs:} updated evaluation history $(\mathcal{H},\mathcal{Y})$ and updated incumbent $(\mathbf{h}_{\mathrm{best}},y_{\mathrm{best}})$.
To guard against subspace stagnation while balancing exploration–exploitation, GIF triggers a full-space step \emph{only} when an entire group-wise round yields no improvement.
Given $T_{\mathrm{left}}$, define the remaining full-space quota $T_{\mathrm{full\,left}} \;=\; \max\!\bigl(0,\; B_{\mathrm{full\,total}} - T_{\mathrm{full\,used}}\bigr),$
and the estimated number of future rounds $n_{\mathrm{round}} \;=\; \left\lfloor \frac{T_{\mathrm{left}}}{B_{\mathrm{step}}} \right\rfloor + 1 .$
Allocate a per-round fallback budget $B_{\mathrm{full}} \;=\; \min\!\Bigl(\,\bigl\lfloor T_{\mathrm{full\,left}}/n_{\mathrm{round}} \bigr\rfloor,\; T_{\mathrm{left}} \Bigr).$
Run the inner optimizer on the full space $\Theta$ for $B_{\mathrm{full}}$ evaluations with warm-start $(\mathcal{H},\mathcal{Y})$, obtain $(\mathcal{H}_{\mathrm{full}},\mathcal{Y}_{\mathrm{full}})$, and update
$(\mathbf{h}_{\mathrm{best}},y_{\mathrm{best}})$,
$T_{\mathrm{used}} \!\leftarrow\! T_{\mathrm{used}} + B_{\mathrm{full}}$,
$T_{\mathrm{full\,used}} \!\leftarrow\! T_{\mathrm{full\,used}} + B_{\mathrm{full}}$. If group-wise optimization keeps improving, the fallback is never activated; the algorithm continues with the standard per-round budget until $T_{\mathrm{used}}=B_{\mathrm{total}}$, and any unused full-space quota remains unspent.

\textbf{Implementation Note}
The inner routine $\mathcal{A}_{\mathrm{opt}}$ can be any standard HPO method (e.g., TPE, BOHB) that supports warm starts. All calls reuse the cumulative history $(\mathcal{H},\mathcal{Y})$, enabling consistent importance estimation and avoiding redundant random initialization. In our experiments, we focus on the scheduling strategy itself and therefore adopt TPE as the default $\mathcal{A}_{\mathrm{opt}}$ unless otherwise specified.

\section{Experiments}
\label{sec:experiments}

We evaluate GIF on three types of benchmarks: (1) anisotropic analytic functions designed to stress high-dimensional search, (2) Bayesmark tabular tasks, and (3) NAS-Bench-301 (33D) for neural architecture optimization. Unless noted otherwise, each run uses a total budget of \(500\) evaluations over \(5\) independent seeds, with the warm-start budget \(B_{\mathrm{init}}{=}100\) counted as part of the total. The inner optimizer \(\mathcal{A}_{\mathrm{opt}}\) is TPE~\cite{akiba2019optuna}, and the importance estimator \(\mathcal{A}_{\mathrm{imp}}\) is N-RReliefF~\cite{wang2024efficient}. For GIF, we use the following default configuration across benchmarks: subsample ratio \(\alpha{=}0.6\), per-round step size \(B_{\mathrm{step}}{=}d\), maximum group size \(k{=}\lfloor d/3 \rfloor\), and fallback ratio \(\rho{=}0.2\).

\subsection{Anisotropic Analytic Function Benchmarks}
\label{subsec:anisotropic_design}

We selected five classic black-box optimization functions widely used in HPO benchmarking: Sphere, Rosenbrock, Ackley~\cite{ackley2012connectionist}, Griewank~\cite{griewank1985generalized}, and Rastrigin~\cite{rastrigin1974systems}. Each function was instantiated at dimensions $d \in \{5,10,30,50\}$. 

\textbf{Anisotropic Variable Transformation.}\label{sec:anisotropic}
To induce anisotropy, we applied a diagonal scaling $\mathbf{w}=(w_1,\dots,w_d)$ with $w_i=\exp\!\big(-\alpha\,(i-1)\big), \alpha=\frac{-\log(10^{-3})}{d-1},$ so that $w_d/w_1=\exp\!\big(-\alpha(d-1)\big)=10^{-3}$. This stylized construction creates a known, non-uniform sensitivity profile across coordinates. It is intended to provide a clean and controlled testbed for evaluating importance-aware schedulers under heterogeneous influence.

\begin{table}[t]
\centering
\setlength{\tabcolsep}{0.5pt}
\renewcommand{\arraystretch}{0.5 }

\caption{Weighted analytic benchmark functions with anisotropic scaling.}
\label{tab:functions}

\begin{tabular}{>{\raggedright\arraybackslash}p{2.9cm} >{\raggedright\arraybackslash}p{5.6cm}}
\toprule
\textbf{Function} & \textbf{Formula} \\
\midrule

Anisotropic Sphere &
$f(\mathbf{x}) = -\sum_{i=1}^d (w_i x_i)^2$ \\

Anisotropic Rosenbrock &
$f(\mathbf{x}) = -\sum_{i=1}^{d-1}\!\Big[100\!\big(w_{i+1}x_{i+1}-(w_i x_i)^2\big)^2 + (1-w_i x_i)^2\Big]$ \\

Anisotropic Ackley &
$\begin{aligned}
f(\mathbf{x}) = {}&-\Big(-20\,\exp\!\big(-0.2\sqrt{\tfrac{1}{d}\sum_{i=1}^d (w_i x_i)^2}\big) \\
&-\exp\!\big(\tfrac{1}{d}\sum_{i=1}^d \cos(2\pi w_i x_i)\big) + 20 + \mathrm{e}\Big)
\end{aligned}$ \\

Anisotropic Griewank &
$f(\mathbf{x}) = -\Big(1 + \tfrac{1}{4000}\sum_{i=1}^d (w_i x_i)^2 - \prod_{i=1}^d \cos\!\big(\tfrac{w_i x_i}{\sqrt{i}}\big)\Big)$ \\

Anisotropic Rastrigin &
$f(\mathbf{x}) = -\sum_{i=1}^d \Big[(w_i x_i)^2 - 10\cos(2\pi w_i x_i) + 10\Big]$ \\

\bottomrule
\end{tabular}
\end{table}

In Table~\ref{tab:functions}, the domains are $[-5,5]^d$ for Sphere, Rosenbrock, Ackley, and Griewank, and $[-5.12,5.12]^d$ for Rastrigin. We negate the standard minimization forms to adopt a maximization convention. For Sphere, Ackley, Griewank, and Rastrigin, the global maximizer is $\mathbf{x}=\mathbf{0}$ with maximum $0$. For Rosenbrock, the \emph{unconstrained} maximizer under our scaling satisfies $w_i x_i=1$ for all $i$, yielding value $0$.

\textbf{Baselines.} We compared GIF against Sequential Grouping (SG)~\cite{wang2025grouped}, Bayesian Optimization based on Tree-structured Parzen Estimator (TPE), Bayesian Optimization based on Gaussian Process (GP), Bayesian Optimization with Hyperband (BOHB)~\cite{falkner2018bohb}, and Random Search. All competitors used the identical box domains in Table~\ref{tab:functions}, the same total evaluation budget (500) and seeds (5), and — where appropriate — the same warm-start history.

\textbf{Verification of Importance Estimation.}\label{sec:verify-importance-analytic} We verified that N\mbox{-}RReliefF could serve as an importance analyzer by testing whether it recovered the coordinate-wise anisotropy of each benchmark function. For every function and each \(d\in\{5,10,30,50\}\), we drew \(500\) i.i.d.\ samples \(\mathbf{x}\sim\mathcal{U}([-1,1]^d)\), evaluated \(y=f(\mathbf{x})\), estimated per-coordinate importances \(\{I_i\}\), and compared them with the ground-truth weights \(\{w_i\}\) after max-normalization. Recovery was quantified by the Pearson correlation between \(\{w_i\}\) and \(\{I_i\}\).

\subsection{Ablation Studies}
\label{subsec:ablation_design}
To isolate the contribution of each design component in GIF, we conducted ablations on the same anisotropic analytic benchmarks as Table~\ref{tab:functions}, using the identical protocol and budgets as in the previous subsection. 
\textbf{Variant A — Randomized Importance (\texttt{RandImp}):} We replaced the importance evaluator with random per–coordinate weights to test whether gains arose from meaningful importance estimation rather than staged optimization alone; 
\textbf{Variant B — Uniform Allocation (\texttt{UniAlloc}):} We retained true importances for grouping but allocated an equal number of trials to each group (no importance weighting) to probe the necessity of importance-weighted budgeting; 
\textbf{Variant C — No Full-Space Fallback (\texttt{NoFB}):} We disabled the joint full-space optimization step to evaluate the fallback’s role in escaping local plateaus and maintaining robustness.

\phantomsection
\label{app:anisotropic_metrics}

For all the experiments on the anisotropic analytic benchmarks, we aggregated across all five functions and $d\in\{5,10,30,50\}$, and reported:
(i) normalized regret AUC (Table~\ref{tab:analytic_results}, the detailed description of the metric is in Sec.~\ref{sec:verify-importance}); and (ii) final best values at 500 trials summarized in a per-function heatmap (mean~$\pm$~std across 5 seeds; Fig.~\ref{fig:analytic_heatmap});

\subsection{Bayesmark}
Bayesmark is an open‐source benchmark for comparing Bayesian optimization methods via a unified API, standardized search spaces, and consistent evaluation~\cite{bayesmark2019}. 
We ran the official Bayesmark benchmark on four datasets (\texttt{breast}, \texttt{digits}, \texttt{iris}, \texttt{wine}) using Bayesmark's default configurations and search spaces for four models---Decision Tree (\texttt{DT}, $d{=}6$), Random Forest (\texttt{RF}, $d{=}6$), Multi-Layer Perceptron trained with the Adam (\texttt{MLP-adam}, $d{=}9$), and Multi-Layer Perceptron trained with stochastic gradient descent (\texttt{MLP-sgd}, $d{=}8$). For consistency across tasks, we used a single primary metric per task type: accuracy for classification and mean squared error (MSE) for regression. Train/validation splits and hyperparameter ranges followed Bayesmark defaults for all optimizers. We summarized performance via task-wise normalized final best scores (Perf.~Norm), Avg.~Rank, Time Rank, and Win Rate in Table~\ref{tab:bayesmark_summary}.

\textbf{Baselines.}\label{subsec:bayesmark_baselines} We included Bayesmark’s default optimizers: \textit{HyperOpt} (HOpt; TPE-based Bayesian optimization), \textit{OpenTuner-BanditA} (OT-B; bandit-coordinated portfolio search), \textit{OpenTuner-GA} (OT-GA; genetic algorithm), \textit{OpenTuner-GA-DE} (OT-GD; GA + differential evolution hybrid), \textit{PySOT} (PySOT; surrogate-based global optimization), \textit{RandomSearch} (RS; uniform random sampling), \textit{Scikit-GBRT-Hedge} (GBRT; GBRT surrogate with Hedge acquisition mixing), \textit{Scikit-GP-Hedge} (GP-H; GP surrogate with Hedge acquisition mixing), and \textit{Scikit-GP-LCB} (GP-LCB; GP surrogate with LCB acquisition)~\cite{bayesmark2019,bergstra2011algorithms,ansel2014opentuner,eriksson2019pysot,head2018scikitopt}. To ensure fairness, all baselines and \textit{GIF} ran with identical search spaces, budgets, seeds, and splits; when applicable, we reused the same warm-start history.

\subsection{NAS-Bench-301}
Unlike the fully tabular NAS benchmarks 101~\cite{ying2019bench} and 201~\cite{dong2020bench}, NAS-Bench-301 (NB301)\cite{zela2020surrogate} is a surrogate benchmark that emulates the Differentiable Architecture Search (DARTS)~\cite{liu2018darts} search space and yields fast, approximate evaluations in a realistic high-dimensional regime. Concretely, NB301 is built on the DARTS cell space trained on CIFAR-10 and provides learned regressors that map an architecture encoding to predicted validation accuracy (and a separate regressor for runtime), enabling faithful anytime comparisons without re-training each architecture. In this work, we used the official \texttt{SNB-DARTS-XGB-v1.0} release~\cite{zela2020surrogate}: an XGBoost-based surrogate trained on DARTS+CIFAR-10 with stratified train/val/test splits over data gathered from multiple NAS optimizers. We kept the benchmark’s 33-dimensional architecture encoding and queried the surrogate-predicted validation accuracy as the objective; for wall-clock plots we used the benchmark’s runtime surrogate to accumulate simulated time. We evaluated GIF against TPE, BOHB, Random, and SG on \texttt{darts-xgb-v1.0}. We reported: (i) best validation score vs.\ evaluations; (ii) best validation score vs.\ simulated wall-clock time; and (iii) a Pareto view (score~vs.~time) that summarizes the quality–time trade-off (Fig.~\ref{fig:nb301_tradeoff}).

\section{Results}
\label{sec:results}

\subsection{Verification of Importance Estimation}
\label{sec:verify-importance}

\begin{table}[t]
\centering
\caption{Pearson correlation between ground-truth weights $w_i$ and HIA score estimates $I_i$.}
\label{tab:pearson-verify}
\setlength{\tabcolsep}{2pt}
\renewcommand{\arraystretch}{1}
\resizebox{\columnwidth}{!}{%
\begin{tabular}{lcccc}
\toprule
 & $d=5$ & $d=10$ & $d=30$ & $d=50$ \\
\midrule
Weighted Ackley     & $0.995$ & $0.986$ & $0.959$ & $0.941$ \\
Weighted Griewank   & $0.985$ & $0.896$ & $0.670$ & $0.547$ \\
Weighted Rastrigin  & $0.990$ & $0.854$ & $0.696$ & $0.717$ \\
Weighted Rosenbrock & $0.993$ & $0.982$ & $0.831$ & $0.791$ \\
Weighted Sphere     & $0.987$ & $0.917$ & $0.819$ & $0.805$ \\
\midrule
Mean $\pm$ Std      & $0.990 \pm 0.004$ & $0.927 \pm 0.057$ & $0.795 \pm 0.116$ & $0.760 \pm 0.144$ \\
\bottomrule
\end{tabular}}
\end{table}

\begin{figure}
\centering
\includegraphics[width=\linewidth, height=4.5cm]{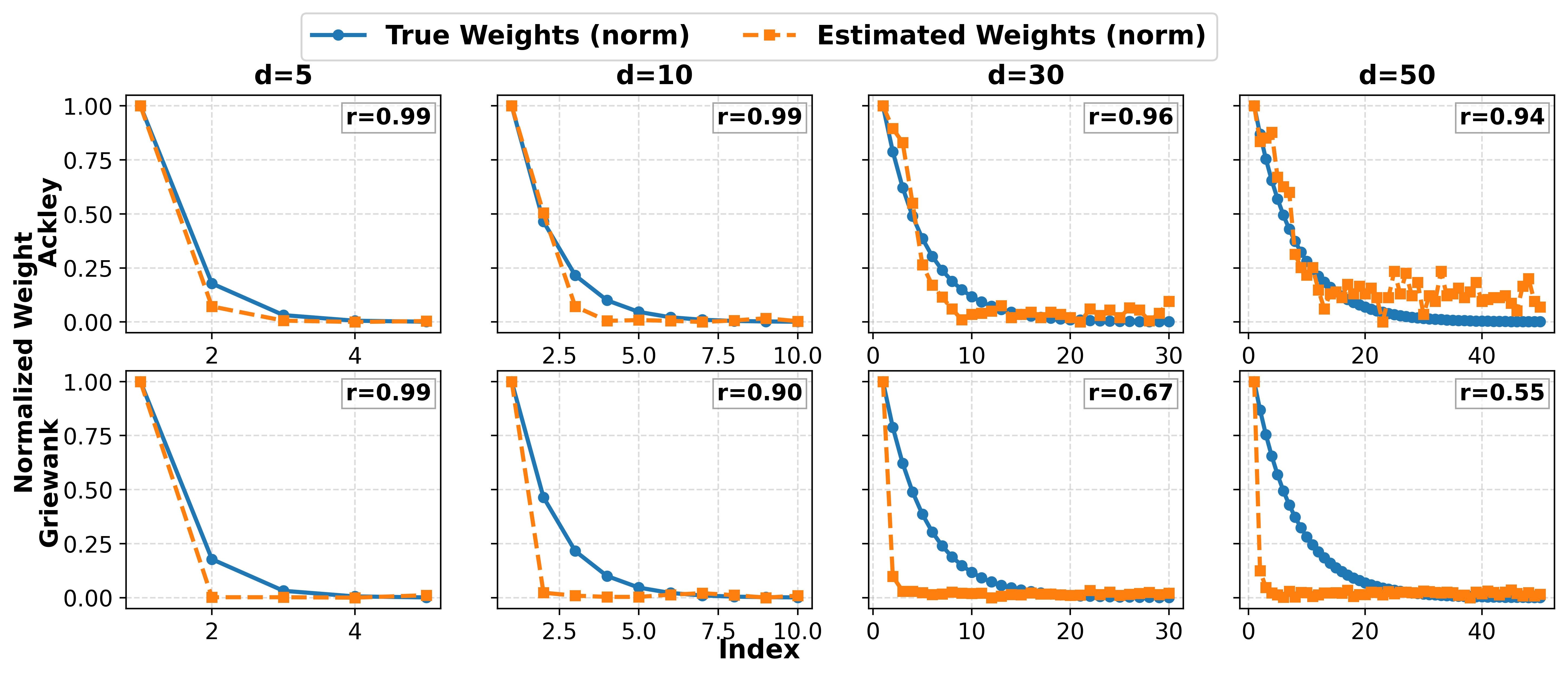}
\caption{Anisotropy verification on weighted Ackley and Griewank: normalized ground-truth weights vs.\ estimated weights across dimensions ($d\in\{5,10,30,50\}$), with Pearson correlation $r$ shown in each subplot.}
\label{fig:ackley-griewank-verify}
\end{figure}

\begin{figure*}[!t]
\centering
\includegraphics[width=0.75\linewidth]{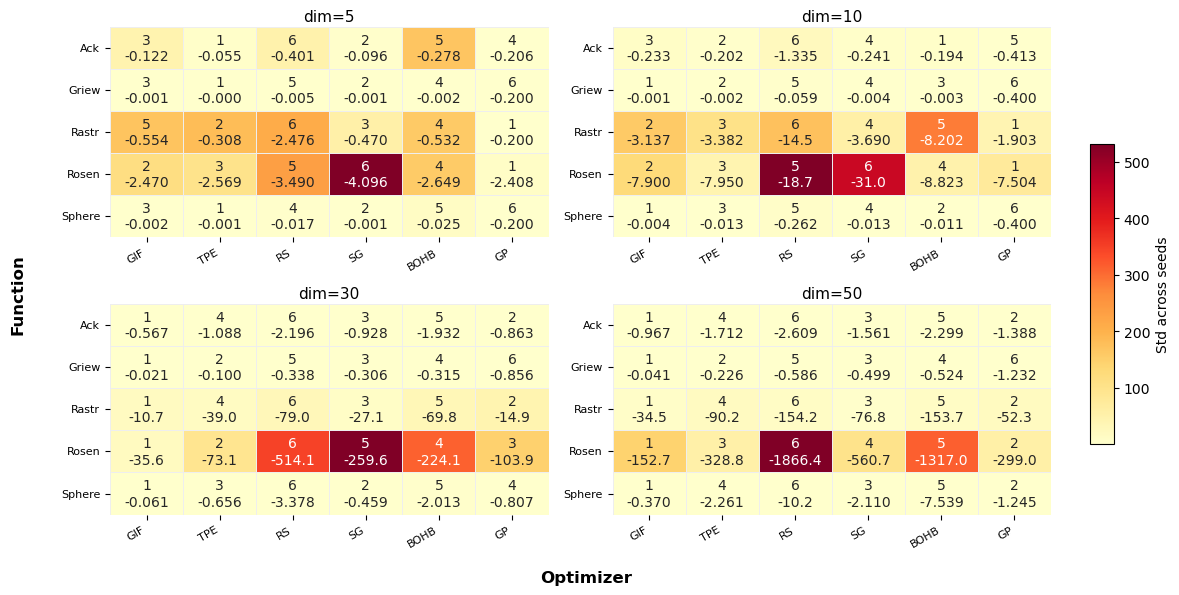}
\caption{
Performance summary of GIF and baselines on weighted analytic benchmarks.
}
\label{fig:analytic_heatmap}
\end{figure*}

Before applying GIF to real HPO tasks, we first test whether the importance estimator (N\mbox{-}RReliefF) can recover the intended anisotropy on the analytic benchmarks. Experimental details are given in Sec.~\ref{subsec:anisotropic_design}. Table~\ref{tab:pearson-verify} reports the Pearson correlation between the estimated scores \(\{I_i\}\) and the ground-truth weights \(\{w_i\}\) for \(d\in\{5,10,30,50\}\). As dimension increases under a fixed sampling budget, the estimates become noisier and the correlations decrease. Figure~\ref{fig:ackley-griewank-verify} shows this trend for two representative functions. On weighted Ackley, the estimated importance profile follows the true decay closely across all dimensions, with only mild degradation as \(d\) grows. Weighted Griewank is noticeably harder: the match deteriorates more quickly, especially in higher dimensions. This is consistent with its oscillatory cosine-product structure, which introduces stronger interactions and makes marginal importance harder to estimate from limited samples. Overall, the results show that the intended anisotropy can be recovered reliably in low and moderate dimensions, and that even in harder high-dimensional cases the estimator still captures the broad decay pattern.

\subsection{Analytic Benchmarks and Ablations}
\label{sec:analytic_ablation}

\begin{table}
\caption{Normalized regret AUC (lower is better) for anisotropic analytic benchmarks. Top: baselines vs.\ GIF. Bottom: ablations. GIF-win = fraction of seeds with the best AUC.}
\label{tab:analytic_results}
\centering
\small
\setlength{\tabcolsep}{2pt}
\renewcommand{\arraystretch}{1}

\resizebox{\columnwidth}{!}{%
\begin{tabular}{c cccccc}
\toprule
\textbf{Dim} & BOHB & GP & RS & SG & TPE & \textbf{GIF-win} \\
\midrule
5  & $4.90\,\pm\,2.07$  & $10.40\,\pm\,3.27$ & $7.01\,\pm\,2.55$  & $9.57\,\pm\,1.36$  & {\bfseries $2.71\,\pm\,0.64$} & 20\% \\
10 & $11.36\,\pm\,3.16$ & $20.35\,\pm\,6.23$ & $21.41\,\pm\,2.72$ & $25.36\,\pm\,5.01$ & $8.06\,\pm\,2.00$             & 60\% \\
30 & $35.09\,\pm\,3.42$ & $30.89\,\pm\,4.52$ & $39.75\,\pm\,6.01$ & $40.14\,\pm\,5.42$ & $21.34\,\pm\,2.84$            & 100\% \\
50 & $44.23\,\pm\,3.57$ & $42.88\,\pm\,2.26$ & $48.36\,\pm\,4.28$ & $47.08\,\pm\,6.15$ & $29.35\,\pm\,2.54$            & 100\% \\
\bottomrule
\end{tabular}}

\vspace{0.35em}

\resizebox{\columnwidth}{!}{%
\begin{tabular}{c ccccc}
\toprule
\textbf{Dim} & \textbf{GIF} & RandImp & UniAlloc & NoFB & \textbf{GIF-win} \\
\midrule
5  & $3.20\,\pm\,0.81$               & $3.24\,\pm\,2.65$ & $3.28\,\pm\,2.63$ & $3.26\,\pm\,2.68$ & 0\% \\
10 & {\bfseries\boldmath $7.36\,\pm\,0.96$} & $6.96\,\pm\,2.55$ & $7.95\,\pm\,2.68$ & $7.19\,\pm\,2.51$ & 60\% \\
30 & {\bfseries\boldmath $14.75\,\pm\,2.55$} & $18.46\,\pm\,3.98$ & $21.01\,\pm\,5.71$ & $17.62\,\pm\,4.39$ & 100\% \\
50 & {\bfseries\boldmath $19.18\,\pm\,1.56$} & $24.07\,\pm\,3.07$ & $24.70\,\pm\,2.97$ & $23.33\,\pm\,2.54$ & 100\% \\
\bottomrule
\end{tabular}}
\vspace{-6pt}
\end{table}
\begin{figure*}
  \centering
  \includegraphics[width=0.9\linewidth]{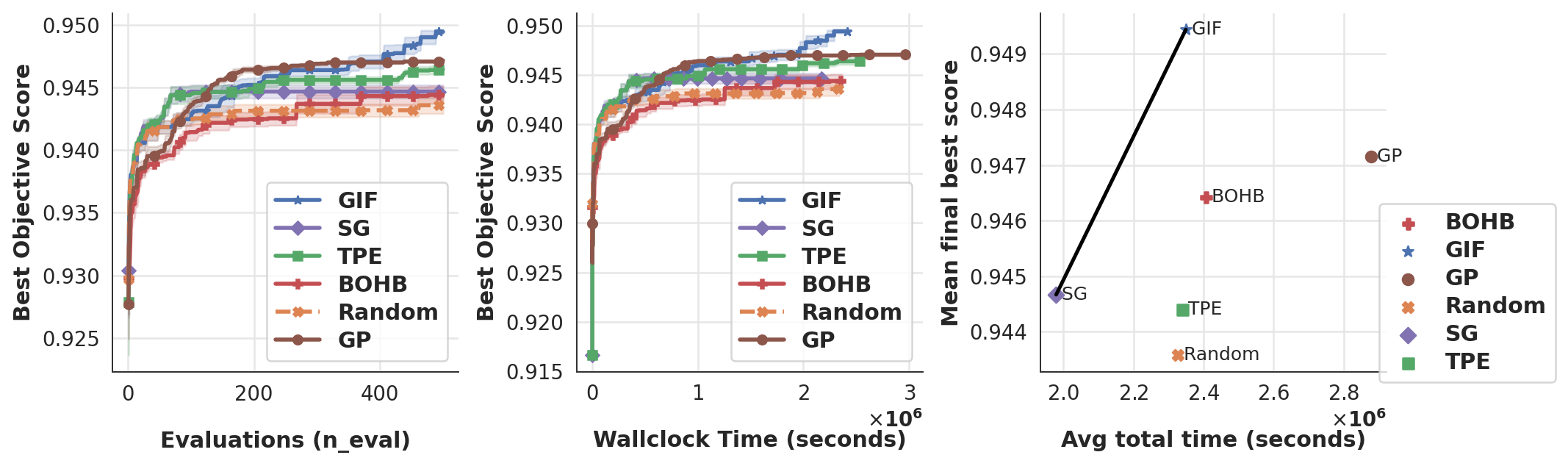}
  \caption{Convergence and Pareto analysis on NAS-Bench-301 (DARTS-XGB surrogate, 33D)}
  \label{fig:nb301_tradeoff}
\end{figure*}

We evaluate GIF on five anisotropic analytic functions against five baselines and three ablation variants. The setup is described in Sec.~\ref{subsec:anisotropic_design} and Sec.~\ref{subsec:ablation_design}. Figure~\ref{fig:analytic_heatmap} shows that at low dimension (\(d{=}5\)), GIF does not consistently dominate strong baselines such as TPE. As dimension increases, however, GIF becomes more reliable across functions, while several baselines degrade or become unstable.

Table~\ref{tab:analytic_results} reports normalized regret AUC (lower is better), following~\cite{klein2017fastbo}. For a maximization objective \(f(\cdot)\), the regret at trial \(t\) is $r_t = f^\star - \max_{s\le t} f\!\left(h^{(s)}\right),$ where \(f^\star\) is the known optimum. We summarize the trajectory over \(T\) trials by $\text{Regret-AUC}=\frac{1}{r_0T}\int_0^T r_t\,dt,$ where \(r_0\) is computed from a shared initialization so that results from different functions are on a comparable scale.

At \(d{=}5\), TPE achieves the best mean score, while GIF is slightly worse and wins only 20\% of seeds. This is not surprising: in small spaces, strong baselines can already model the landscape well, so the benefit of importance-aware scheduling is limited. From \(d{=}10\) onward, GIF consistently performs better than all baselines. Its win rate also increases sharply with dimension, suggesting that the gains are not only larger on average but also more stable across seeds.

The ablation results show that each component matters. Replacing learned importance with random weights (\texttt{RandImp}) or removing proportional allocation (\texttt{UniAlloc}) leads to worse AUC, indicating that the importance signal itself is useful. Removing the fallback step (\texttt{NoFB}) is especially harmful in higher dimensions, where periodic global exploration helps recover from poor subspace choices.

\subsection{Bayesmark (Mid-Dimensional Evaluation)}
\label{sec:bayesmark}
\begin{table}
\caption{Bayesmark summary. Optimizer abbreviations are defined in Sec.~\ref{subsec:bayesmark_baselines}.}
\label{tab:bayesmark_summary}
\centering
\resizebox{\columnwidth}{!}{%
\begin{tabular}{lcccc}
\toprule
Opt. & Avg.~Rank $\downarrow$ & Time Rank $\downarrow$ & Perf.~Norm $\uparrow$ & Win Rate $\uparrow$ \\
\midrule
GIF    & \textbf{2.72} & 5.13  & \textbf{0.811} & \textbf{0.750} \\
RS     & 6.28          & \textbf{1.66} & 0.189 & 0.063 \\
HOpt   & 4.56          & 5.84  & 0.354 & 0.094 \\
PySOT  & 4.47          & 6.03  & 0.371 & 0.156 \\
GP-H   & 3.47          & 10.0  & 0.401 & 0.125 \\
OT-B   & 5.28          & 3.78  & 0.274 & 0.094 \\
GBRT   & 4.78          & 7.94  & 0.314 & 0.125 \\
OT-GD  & 5.50          & 2.94  & 0.234 & 0.063 \\
GP-LCB & 4.09          & 9.00  & 0.370 & 0.219 \\
OT-GA  & 5.75          & 2.69  & 0.220 & 0.063 \\
\bottomrule
\end{tabular}}
\end{table}

Having verified the importance estimator and tested GIF on controlled analytic functions, we next evaluate it on Bayesmark. Table~\ref{tab:bayesmark_summary} summarizes performance across tasks using four metrics: mean normalized final score (Perf.~Norm), average rank by performance, average rank by runtime, and win rate.

GIF is the strongest method overall in this benchmark: it achieves the best Perf.~Norm, the best average rank, and the highest win rate. Random Search remains attractive when runtime is the main concern, but its final performance is much weaker. Other surrogate-based methods, such as HOpt and PySOT, can perform well on some tasks, but they are less consistent across the full benchmark.

\begin{figure}
\centering
\includegraphics[width=\linewidth]{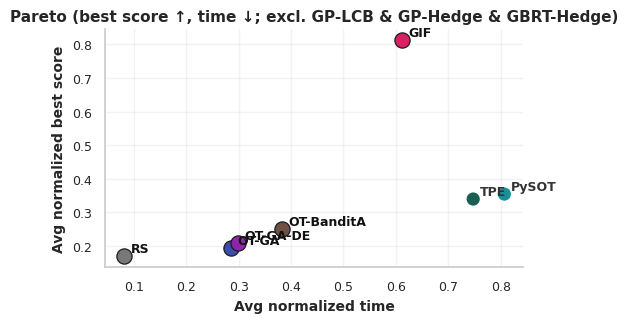}
\caption{Pareto trade-off between final score and time (lightweight methods).}
\label{fig:bayesmark_pareto}
\end{figure}

This result is in line with the role of GIF. Bayesmark contains tasks with different datasets, models, and response surfaces, so performance on one subset of tasks does not necessarily transfer to the others. GIF is not the fastest method because importance estimation, grouping, and fallback steps introduce extra overhead. Still, that extra coordination improves how evaluations are spent, which leads to better final solutions overall. The Pareto plot in Fig.~\ref{fig:bayesmark_pareto} reflects this trade-off: GIF lies on the frontier, offering stronger final quality at a moderate time cost.

\subsection{NasBench 301 (High-Dimensional Evaluation)}

We now shift to a genuinely high-dimensional setting: NAS-Bench-301 (33D, DARTS-XGB surrogate). Figure~\ref{fig:nb301_tradeoff} summarizes convergence in evaluations (left), wall-clock time (middle), and the score–time Pareto view (right). In evaluations, GIF keeps improving after other methods flatten out; around 340 evaluations, it overtakes all baselines. This indicates that the importance-guided scheduler continues to discover productive subspaces late in the run, and the warm-started full-space fallback helps it escape plateaus. In wall-clock time, GIF uses the budget efficiently: it reaches the top accuracy without being the slowest; SG is faster but stalls at a lower ceiling, and GP is slowest at the same budget. The Pareto panel makes the trade-off explicit: GIF and SG define the frontier—SG at the “faster but lower score” end, GIF at the “higher score at similar time” end—while GP, TPE, BOHB, and Random are dominated (either slower for similar accuracy or less accurate at similar time). As a result, in high dimensions, focusing trials on the most important groups while retaining periodic full-space search yields stronger final incumbents and higher accuracy for the time spent.

\section{Conclusion}
\label{sec:conclusion}
Our study introduces \textsc{Greedy Importance First} (GIF), an importance-aware strategy that translates early hyperparameter-importance estimates into concrete scheduling decisions—grouping by importance, proportional allocation, and a safeguarded full-space fallback. Across diverse benchmarks, a consistent pattern emerges: GIF is most effective in high-dimensional regimes. On weighted analytic functions and NAS-Bench-301, it achieves both faster convergence and stronger final incumbents than strong baselines. On Bayesmark, where the effective dimensionality is smaller, GIF remains competitive, but its margins are limited on simpler models and become most pronounced on the MLPs—reflecting that importance-guided scheduling yields the biggest gains when many low-impact variables dilute progress and the landscape exhibits stronger anisotropy. In summary, GIF provides a simple, plug-compatible pathway to sample-efficient HPO in high dimensions; by reweighting effort toward important subspaces while maintaining a robust fallback, it offers practical utility for deep learning model tuning, and lays a foundation for future research on importance-aware AutoML systems.

\section*{Acknowledgment}
RW acknowledges support from the China Scholarship Council.
MG acknowledges support from the EPSRC project EP/X001091/1.

\bibliographystyle{IEEEtran}
\bibliography{references}

@inproceedings{falkner2018bohb,
  title={BOHB: Robust and efficient hyperparameter optimization at scale},
  author={Falkner, Stefan and Klein, Aaron and Hutter, Frank},
  booktitle={International Conference on Machine Learning},
  pages={1437--1446},
  year={2018},
  organization={PMLR}
}

@article{li2018hyperband,
  title={Hyperband: A novel bandit-based approach to hyperparameter optimization},
  author={Li, Lisha and Jamieson, Kevin and DeSalvo, Giulia and Rostamizadeh, Afshin and Talwalkar, Ameet},
  journal={Journal of Machine Learning Research},
  volume={18},
  number={185},
  pages={1--52},
  year={2018}
}

@article{probst2019tunability,
  title={Tunability: Importance of hyperparameters of machine learning algorithms},
  author={Probst, Philipp and Boulesteix, Anne-Laure and Bischl, Bernd},
  journal={Journal of Machine Learning Research},
  volume={20},
  number={53},
  pages={1--32},
  year={2019}
}

@article{bischl2023hyperparameter,
  title={Hyperparameter optimization: Foundations, algorithms, best practices, and open challenges},
  author={Bischl, Bernd and Binder, Martin and Lang, Michel and Pielok, Tobias and Richter, Jakob and Coors, Stefan and Thomas, Janek and Ullmann, Theresa and Becker, Marc and Boulesteix, Anne-Laure and others},
  journal={Wiley Interdisciplinary Reviews: Data Mining and Knowledge Discovery},
  volume={13},
  number={2},
  pages={e1484},
  year={2023},
  publisher={Wiley Online Library}
}

@article{bergstra2011algorithms,
  title={Algorithms for hyper-parameter optimization},
  author={Bergstra, James and Bardenet, R{\'e}mi and Bengio, Yoshua and K{\'e}gl, Bal{\'a}zs},
  journal={Advances in neural information processing systems},
  volume={24},
  year={2011}
}

@inproceedings{hutter2014efficient,
  title={An efficient approach for assessing hyperparameter importance},
  author={Hutter, Frank and Hoos, Holger and Leyton-Brown, Kevin},
  booktitle={International conference on machine learning},
  pages={754--762},
  year={2014},
  organization={PMLR}
}

@inproceedings{wang2024efficient,
  title={Efficient hyperparameter importance assessment for cnns},
  author={Wang, Ruinan and Nabney, Ian and Golbabaee, Mohammad},
  booktitle={International Conference on Neural Information Processing},
  pages={16--31},
  year={2024},
  organization={Springer}
}

@article{watanabe2023ped,
  title={PED-ANOVA: efficiently quantifying hyperparameter importance in arbitrary subspaces},
  author={Watanabe, Shuhei and Bansal, Archit and Hutter, Frank},
  journal={arXiv preprint arXiv:2304.10255},
  year={2023}
}

@misc{bayesmark2019,
  title        = {Bayesmark: Benchmark framework to compare Bayesian optimization methods on real ML tasks},
  author       = {Turner, Ryan D. and Eriksson, David},
  year         = {2019},
  howpublished = {\url{https://github.com/uber/bayesmark}},
  note         = {Accessed 2025-09-14}
}

@article{loshchilov2016cma,
  title={CMA-ES for hyperparameter optimization of deep neural networks},
  author={Loshchilov, Ilya and Hutter, Frank},
  journal={arXiv preprint arXiv:1604.07269},
  year={2016}
}

@article{zela2020surrogate,
  title={Surrogate NAS benchmarks: Going beyond the limited search spaces of tabular NAS benchmarks},
  author={Zela, Arber and Siems, Julien and Zimmer, Lucas and Lukasik, Jovita and Keuper, Margret and Hutter, Frank},
  journal={arXiv preprint arXiv:2008.09777},
  year={2020}
}

@misc{wandb_sweeps,
  author       = {{Weights \& Biases}},
  title        = {Visualize Sweep Results},
  howpublished = {\url{https://docs.wandb.ai/guides/sweeps/visualize-sweep-results/}},
  note         = {Accessed: 2025-09-18},
  year         = {2025}
}

@article{lindauer2022smac3,
  title={SMAC3: A versatile Bayesian optimization package for hyperparameter optimization},
  author={Lindauer, Marius and Eggensperger, Katharina and Feurer, Matthias and Biedenkapp, Andr{\'e} and Deng, Difan and Benjamins, Carolin and Ruhkopf, Tim and Sass, Ren{\'e} and Hutter, Frank},
  journal={Journal of Machine Learning Research},
  volume={23},
  number={54},
  pages={1--9},
  year={2022}
}

@inproceedings{akiba2019optuna,
  title={Optuna: A next-generation hyperparameter optimization framework},
  author={Akiba, Takuya and Sano, Shotaro and Yanase, Toshihiko and Ohta, Takeru and Koyama, Masanori},
  booktitle={Proceedings of the 25th ACM SIGKDD international conference on knowledge discovery \& data mining},
  pages={2623--2631},
  year={2019}
}

@article{liu2024uq,
  title={UQ-guided hyperparameter optimization for iterative learners},
  author={Liu, Jiesong and Zhang, Feng and Guan, Jiawei and Shen, Xipeng},
  journal={Advances in Neural Information Processing Systems},
  volume={37},
  pages={386--415},
  year={2024}
}

@inproceedings{mehta2024improving,
  title={Improving hyperparameter optimization with checkpointed model weights},
  author={Mehta, Nikhil and Lorraine, Jonathan and Masson, Steve and Arunachalam, Ramanathan and Bhat, Zaid Pervaiz and Lucas, James and Zachariah, Arun George},
  booktitle={European Conference on Computer Vision},
  pages={75--96},
  year={2024},
  organization={Springer}
}

@article{wistuba2018scalable,
  title={Scalable gaussian process-based transfer surrogates for hyperparameter optimization},
  author={Wistuba, Martin and Schilling, Nicolas and Schmidt-Thieme, Lars},
  journal={Machine Learning},
  volume={107},
  number={1},
  pages={43--78},
  year={2018},
  publisher={Springer}
}

@article{swersky2013multi,
  title={Multi-task bayesian optimization},
  author={Swersky, Kevin and Snoek, Jasper and Adams, Ryan P},
  journal={Advances in neural information processing systems},
  volume={26},
  year={2013}
}

@article{wang2025grouped,
  title={Grouped Sequential Optimization Strategy--the Application of Hyperparameter Importance Assessment in Deep Learning},
  author={Wang, Ruinan and Nabney, Ian and Golbabaee, Mohammad},
  journal={arXiv preprint arXiv:2503.05106},
  year={2025}
}

@article{liu2018darts,
  title={Darts: Differentiable architecture search},
  author={Liu, Hanxiao and Simonyan, Karen and Yang, Yiming},
  journal={arXiv preprint arXiv:1806.09055},
  year={2018}
}

@inproceedings{ying2019bench,
  title={Nas-bench-101: Towards reproducible neural architecture search},
  author={Ying, Chris and Klein, Aaron and Christiansen, Eric and Real, Esteban and Murphy, Kevin and Hutter, Frank},
  booktitle={International conference on machine learning},
  pages={7105--7114},
  year={2019},
  organization={PMLR}
}

@article{dong2020bench,
  title={Nas-bench-201: Extending the scope of reproducible neural architecture search},
  author={Dong, Xuanyi and Yang, Yi},
  journal={arXiv preprint arXiv:2001.00326},
  year={2020}
}

@book{ackley2012connectionist,
  title={A connectionist machine for genetic hillclimbing},
  author={Ackley, David},
  volume={28},
  year={2012},
  publisher={Springer science \& business media}
}

@article{griewank1985generalized,
  title={Generalized descent for global optimization},
  author={Griewank, AO},
  journal={JOTA},
  volume={34},
  pages={15},
  year={1985}
}

@article{rastrigin1974systems,
  title={Systems of extremal control},
  author={Rastrigin, Leonard Andreevi{\v{c}}},
  journal={Nauka},
  year={1974}
}

@inproceedings{ansel2014opentuner,
  title={Opentuner: An extensible framework for program autotuning},
  author={Ansel, Jason and others},
  booktitle={International Conference on Parallel Architectures and Compilation Techniques},
  pages={303--315},
  year={2014},
  organization={IEEE}
}

@inproceedings{eriksson2019pysot,
  title={Scalable Global Optimization via Local Bayesian Optimization},
  author={Eriksson, David and Bindel, David and Shoemaker, Christine A},
  booktitle={Advances in Neural Information Processing Systems},
  year={2019}
}

@misc{head2018scikitopt,
  title={Scikit-optimize/scikit-optimize},
  author={Head, Tim and MechCoder and Louppe, Gilles and Shcherbatyi, Iaroslav and Fokoue, Ernest and others},
  year={2018},
  howpublished={\url{https://scikit-optimize.github.io/}}
}

@inproceedings{klein2017fastbo,
  title     = {Fast Bayesian Optimization of Machine Learning Hyperparameters on Large Datasets},
  author    = {Klein, Aaron and Falkner, Stefan and Springenberg, Jost Tobias and Hutter, Frank},
  booktitle = {Proceedings of the 20th International Conference on Artificial Intelligence and Statistics (AISTATS)},
  year      = {2017}
}

\end{document}